\documentclass[runningheads]{src/llncs}

\usepackage{graphicx}
\usepackage{src/comment}
\usepackage{amsmath,amssymb} \usepackage{color}
\usepackage{xspace}
\usepackage{breakurl}
\PassOptionsToPackage{hyphens}{url}\usepackage{hyperref}

\usepackage{colortbl}
\usepackage{booktabs}
\usepackage{enumitem}

\newcommand{\sysname}{\textsc{Point\-AR}\xspace}

\definecolor{pro_green}{rgb}{0.0, 0.66, 0.47}
\definecolor{overleaf_green}{rgb}{0.08, 0.54, 0.02}

\newcommand{\ctsong}{\text{Song et al.}~\cite{Song2019}\xspace}
\newcommand{\ctgaron}{\text{Garon et al.}~\cite{Garon2019}\xspace}

\newcommand{\ctpointconv}{\text{Wu et al.}~\cite{Wu_2019_CVPR}\xspace}

\newcommand{\1}{{\em (i)}}
\newcommand{\2}{{\em (ii)}}
\newcommand{\3}{{\em (iii)}}
\newcommand{\4}{{\em (iv)}}

\newcommand{\sh}{spherical harmonics\xspace}
\newcommand{\shc}{spherical harmonics coefficients\xspace}
\newcommand{\SHc}{SH coefficients\xspace}
\newcommand{\mci}{Monte Carlo Integration\xspace}

\begin{document}
\pagestyle{headings}
\mainmatter

\title{\sysname: Efficient Lighting Estimation for Mobile Augmented Reality}

\author{Yiqin Zhao \and
Tian Guo}
\authorrunning{Y. Zhao and T. Guo}
\institute{Worcester Polytechnic Institute \\
\email{\{yzhao11, tian\}@wpi.edu}
}
\maketitle

\begin{abstract}
We propose an efficient lighting estimation pipeline that is suitable to run on modern mobile devices, with comparable resource complexities to state-of-the-art mobile deep learning models.
Our pipeline, \sysname, takes a single RGB-D image captured from the mobile camera and a 2D location in that image, and estimates 2nd order \shc.
This estimated \shc can be directly utilized by rendering engines for supporting spatially variant indoor lighting, in the context of augmented reality.
Our key insight is to formulate the lighting estimation as a point cloud-based learning problem directly from point clouds, which is in part inspired by the Monte Carlo integration leveraged by real-time \sh lighting.
While existing approaches estimate lighting information with complex deep learning pipelines, our method focuses on reducing the computational complexity.
Through both quantitative and qualitative experiments, we demonstrate that \sysname achieves lower lighting estimation errors compared to state-of-the-art methods. Further, our method requires an order of magnitude lower resource, comparable to that of mobile-specific DNNs.

\end{abstract}

\section{Introduction}
\label{sec:intro}

In this paper, we describe the problem of lighting estimation in the context of mobile augmented reality (AR) applications for indoor scene.
We focus on recovering scene lighting, from a partial view, to a representation within the widely used image-based lighting model~\cite{debevec2006image}.
Accurate lighting estimation positively impacts realistic rendering, making it an important task in real-world mobile AR scenarios, e.g., furniture shopping apps that allow user to place a chosen piece in a physical environment.   

In the image-based lighting model, to \emph{obtain} the lighting information at a given position in the physical environment, one would use a $360^{\circ}$ panoramic camera that can capture incoming lighting from every direction. 
However, commodity mobile phones often lack such panoramic camera, making it challenging to directly obtain accurate lighting information and necessitating the task of.
There are three key challenges when estimating lighting information for mobile AR applications.
First, the AR application needs to estimate the lighting at the rendering location, i.e., where the 3D object will be placed, from the camera view captured by the mobile device.
Second, as the mobile camera often only has a limited field of view (FoV), i.e., less than 360 degrees, the AR application needs to derive or estimate the lighting information outside the FoV.
Lastly, as lighting information is used for rendering, the estimation should be fast enough and ideally to match the frame rate of 3D object rendering.

\begin{figure}[t]
\centering
\includegraphics[width=0.95\textwidth]{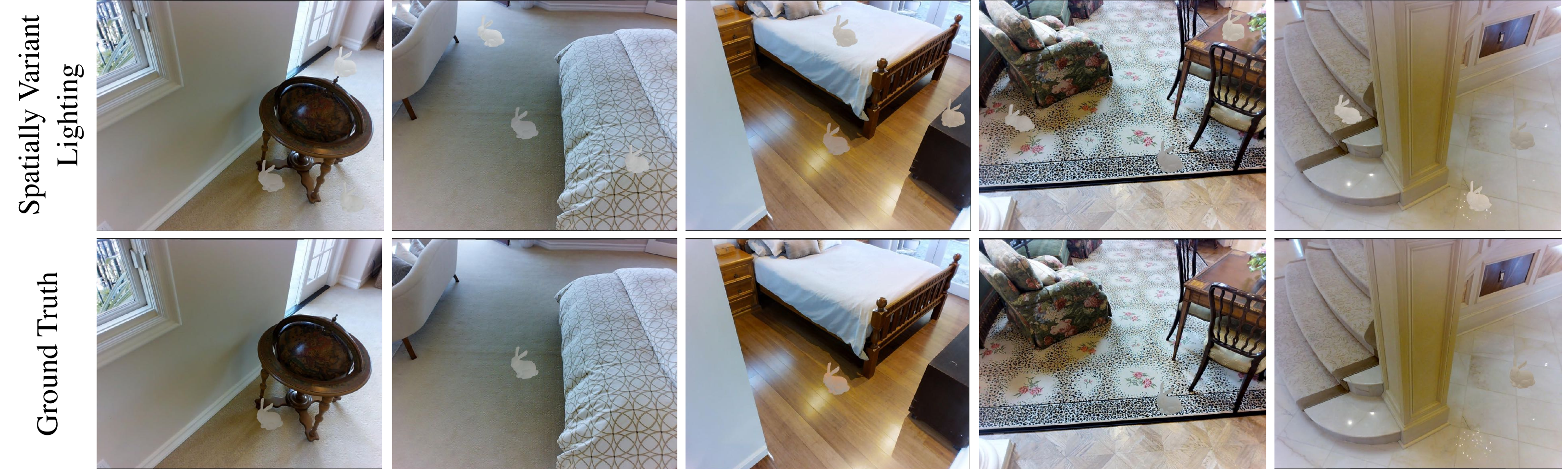}
\caption{\textbf{Rendering examples of \sysname in spatially variant lighting conditions}.
Row 1 shows the Stanford bunnies that were lit using the \shc predicted by \sysname. 
Row 2 shows the ground truth rendering results.
}
\label{fig:spatial_variant}
\end{figure}

Recently proposed learning-based lighting estimation approaches~\cite{Gardner2017,Garon2019,Song2019} did not consider the aforementioned unique challenges of supporting lighting estimation for Mobile AR.
Gardner et al. proposed a simple transformation to tackle the spatial difference between observation and rendering positions~\cite{Gardner2017}. However, the proposed transformation did not use the depth information and therefore can lead to image distortion.
Garon et al. improved the spatial lighting estimations with a two-branches neural network that was reported to perform well on a laptop GPU but not on mobile devices~\cite{Garon2019}.
Song et al. further improved the estimation accuracy by decomposing the pipeline into differentiable sub-tasks~\cite{Song2019}. However, the overall network is large in size and has high computational complexity, which makes it ill-suited for running on mobile phones.

Our key insight is to break down the lighting estimation problem into two sub-problems: \1 geometry-aware view transformation and 
\2 point-cloud based learning from limited scene. 
At a high level, geometry-aware view transformation handles the task of applying spatial transformation to a camera view with a mathematical model. 
In other words, we skip the use of neural networks for considering scene geometry, unlike previous methods that approached the lighting estimation with a monolithic network~\cite{Gardner2017,Garon2019}.
Stripping down the complexity of lighting estimation is crucial as it makes designing mobile-oriented learning models possible. Our key idea for learning lighting information directly from point clouds, instead of images, is in part inspired by the use of \mci in the real-time \sh lighting calculation. 

Concretely, we propose a mobile-friendly lighting estimation pipeline \sysname that combines both physical knowledge and neural network.
We rethink and redefine the lighting estimation pipeline by leveraging an efficient mathematical model to tackle the view transformation and a compact deep learning model for point cloud-based lighting estimation. Our two-stage lighting estimation for mobile AR has the promise of realistic rendering effects and fast estimation speed.

\sysname takes the input of an RGB-D image and a 2D pixel coordinate (i.e., observation position) and outputs the 2nd order spherical harmonics (SH) coefficients (i.e., a compact lighting representation of diffuse irradiance map) at a world position. The estimated \SHc can be directly used for rendering 3D objects, even under spatially variant lighting conditions.
Figure~\ref{fig:spatial_variant} shows the visually satisfying rendering effects of Stanford bunnies at three locations with different lighting conditions.
In summary, \sysname circumvents the hardware limitation (i.e., 360 degree cameras) and enables fast lighting estimation on commodity mobile phones. 

We evaluated our method by training on a point cloud dataset generated from large-scale real-world datasets called Matterport3D and the one from Neural Illumination~\cite{chang2017matterport3d,Song2019}. Compared to recently proposed lighting estimation approaches, our method \sysname achieved up to $31.3\%$ better irradiance map $l2$ loss with one order of magnitude smaller and faster model. Further, \sysname produces comparable rendering effects to ground truth and has the promise to run efficiently on commodity mobile devices.

\section{Mobile AR Lighting Estimation and Its Challenges}

\begin{figure}[t]
\centering
\includegraphics[width=0.95\textwidth]{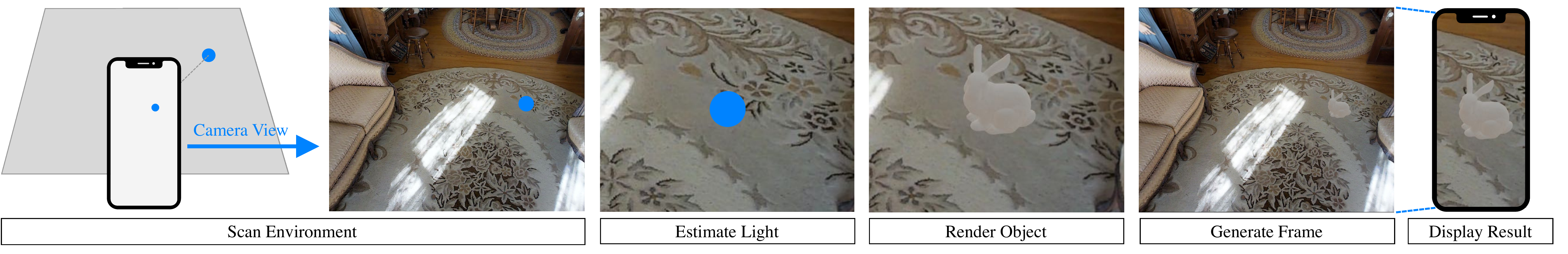}
\caption{\textbf{Lighting estimation workflow in AR applications.}
Lighting estimation starts from a camera view captured by a user's mobile phone camera.
The captured photo, together with a screen coordinate (e.g., provided by the user through touchscreen), is then passed to a lighting estimation algorithm.
The estimated lighting information is then used to render 3D objects, which are then combined with the original camera view into a 2D image frame.
}
\label{fig:arflow}
\end{figure} 

We describe how lighting estimation, i.e., recovering scene lighting based on limited scene information, fits into the mobile augmented workflow from an end-user's perspective. 
The description here focuses on how lighting estimation component will be used while a human user is interacting with a mobile AR application such as a furniture shopping app.
Understanding the role played by lighting estimation module can underscore the challenges and inform the design principles of mobile lighting estimation.

Figure~\ref{fig:arflow} shows how a mobile user with a multi-cameras mobile phone, such as iPhone 11 or Samsung S10, interacts with the mobile AR application. 
Such mobile devices are increasingly popular and can capture image in RGB-D format, i.e., with depth information. 
The user can tap the mobile screen to place the 3D virtual object, such as a couch, on the detected surface. 
To achieve a realistic rendering of the 3D object, i.e., seamlessly blending to the physical environment, the mobile device leverages lighting estimation methods such as those provided by AR frameworks~\cite{arcore_website,arkit_website}. The estimated lighting information will then be used by the rendering engine to relit the 3D object. 

In this work, we target estimating indoor lighting which can change both spatially, e.g., due to user movement, and temporally, e.g., due to additional light sources. To provide good end-user experiences, the mobile AR application often needs to re-estimate lighting information in a rate that matches desired fresh rate measured in frame per second (fps). This calls for fast lighting estimation method that can finish execute in less than 33ms (assuming 30fps). 

However, lighting estimation for mobile AR comes with three inherent challenges that might benefit from deep learning approaches~\cite{Gardner_2019_ICCV,Gardner2017,Garon2019,Song2019}. First, obtaining accurate lighting information for mobile devices is challenging as it requires access to the $360^\circ$ panorama of the rendering position; mobile devices at best can obtain the lighting information at the device location, also referred to as \emph{observation position}, through the use of ambient light sensor or both front- and rear-facing cameras to expand limited field-of-view (FoV).
Second, as indoor lighting often varies spatially, directly using lighting information at the observation location to render a 3D object can lead to undesirable visual effect.
Third, battery-powered commodity mobile devices, targeted by our work, have limited hardware supports when comparing to specialized devices such as Microsoft HoloLens. This further complicates the network designs and emphasizes the importance of mobile resource efficiency. 
 
\section{Problem Formulation}

We formulate the \emph{lighting estimation in mobile augmented reality} as a \SHc regression problem as $h: h(g (f (C, D, I), r)) = S_r$ where $g: g(P_{o}, r) = P_{r}$ and $f$: $f(C, D, I) = P$.
Specifically, we decompose the problem into two stages to achieve the goal of fast lighting estimation on commodity mobile phones. 

The first stage starts with an operation $f(C, D, I)$ that generates a point cloud $P_{o}$ at observation position $o$. This operation takes three inputs: \1 an RGB image, represented as $C$, \2 the corresponding depth image, represented as $D$; and \3 the mobile camera intrinsic $I$. Then $g(P_{o}, r)$ takes both $P_{o}$ and the rendering position $r$, and leverages a linear translation $T$ to generate a point cloud $P_r$ centered at $r$. In essence, this transformation simulates the process of re-centering the camera from user's current position $o$ to the rendering position $r$. We describe the geometric-aware transformation design in Section~\ref{subsec:point_cloud_gen}.

For the second stage, we formulate the lighting estimation as a point cloud based learning problem $h$ that takes an incomplete point cloud $P_r$ and outputs 2nd order \SHc $S_r$. We describe the end-to-end point cloud learning in Sections~\ref{subsec:shc_estimation} to Section~\ref{subsec:discussion}.

\section{\sysname Pipeline Architecture}
\label{sec:pipeline}

\begin{figure}[t]
\centering
\includegraphics[width=0.9\textwidth]{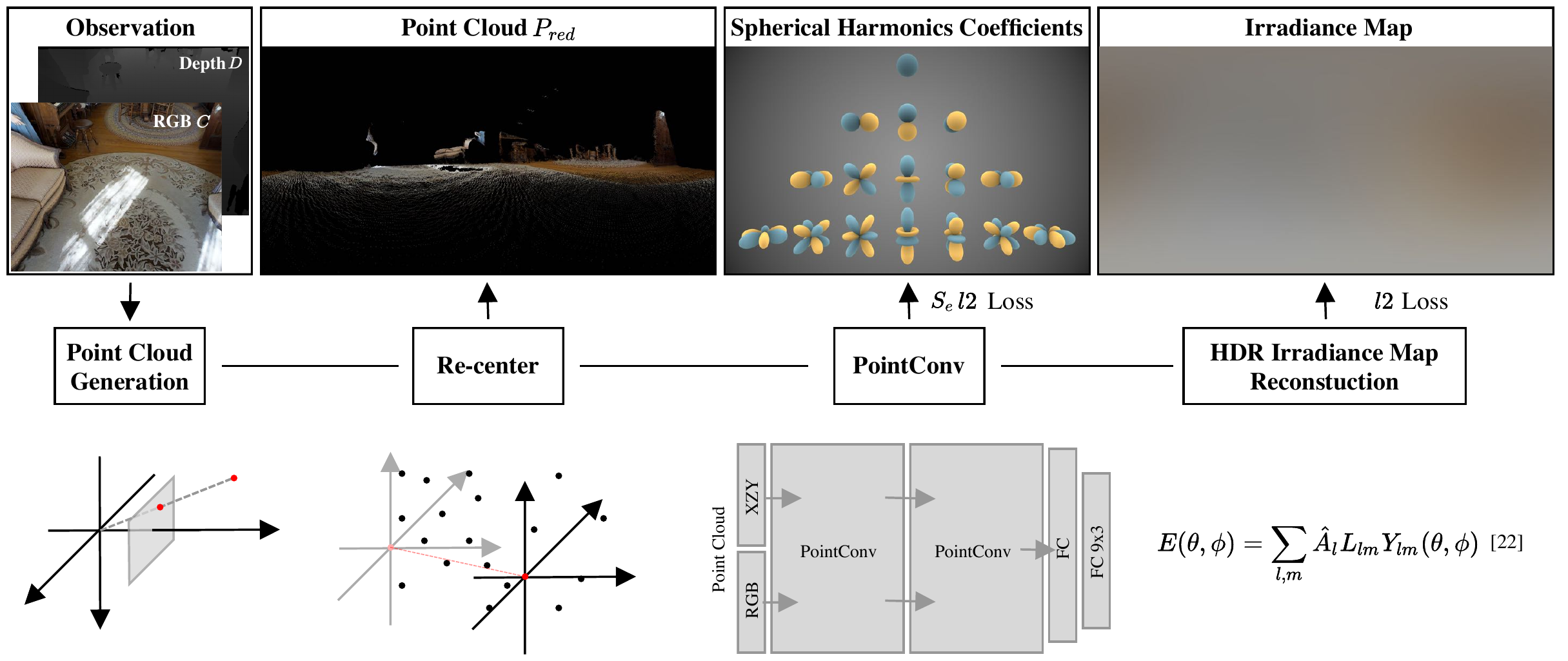}
\caption{\textbf{\sysname pipeline composition and components.} 
We first transform the camera view into a point cloud centered at the observation position, as described in Section~\ref{subsec:point_cloud_gen}.
Then we use a compact neural network described in Section~\ref{subsec:shc_estimation} to estimate \SHc at the rendering position. 
We use $l2$ loss on both estimated \SHc and HDR irradiance map reconstructed from \shc for evaluation.
}
\label{fig:pipeline}
\end{figure}

In this section, we describe our two-stage lighting estimation pipeline \sysname, as shown in Figure~\ref{fig:pipeline}, that is used to model $h$. The first stage includes a point cloud generation and a geometry transformation modules (Section~\ref{subsec:point_cloud_gen}). The second stage corresponds to a deep learning model for estimating lighting information, represented as \SHc (Section~\ref{subsec:shc_estimation}). 
Compared to traditional end-to-end neural network designs~\cite{Gardner2017,Garon2019,Song2019}, \sysname is more resource efficient (as illustrated in Section~\ref{sec:eval}) and exhibits better interpretability. We describe our dataset generation in Section~\ref{subsec:dataset_generation} and our design rationale in Section~\ref{subsec:discussion}.

\begin{figure}[t]
\centering
\includegraphics[width=0.95\textwidth]{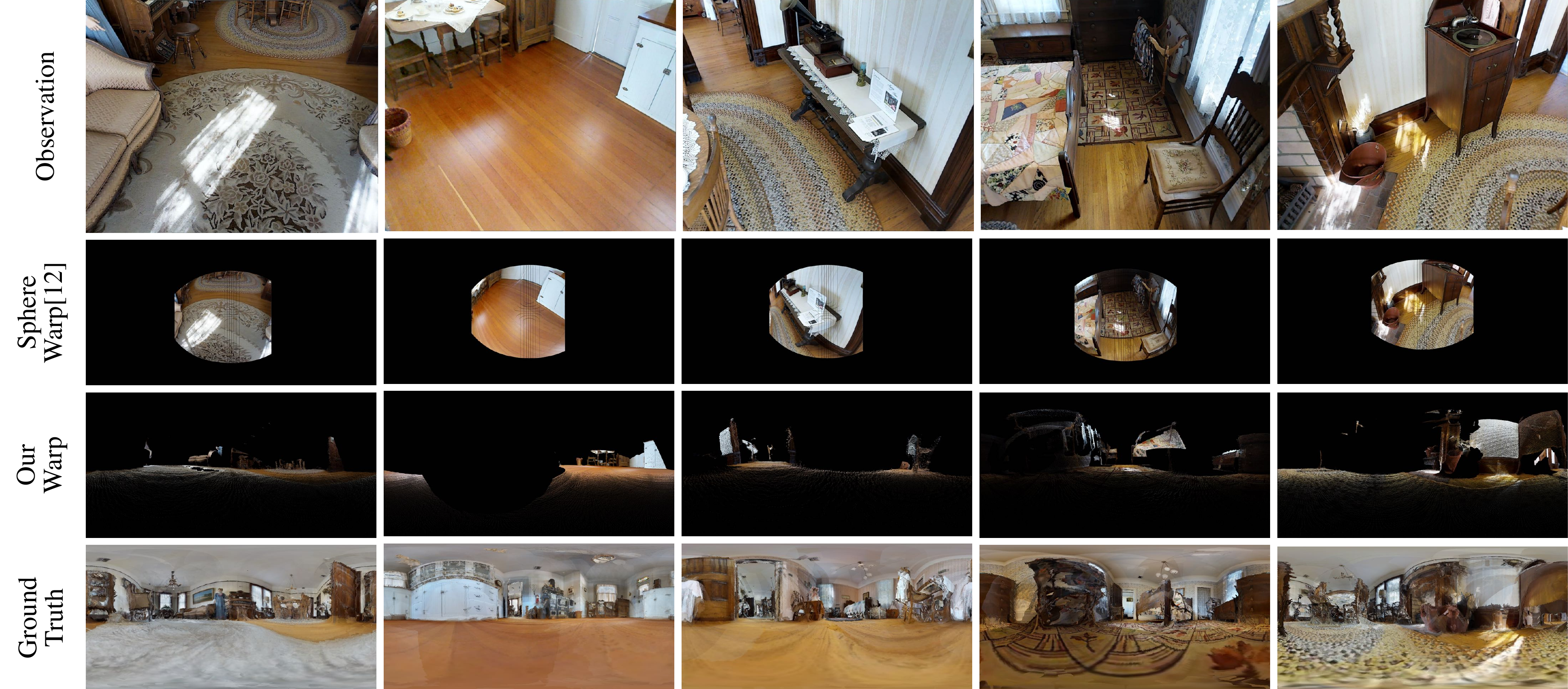}
\caption{\textbf{Comparison of different warping methods.}
To transform mobile camera view spatially, i.e., from observation to rendering location, we leverage the point cloud which was generated from the RGB-D image. Our approach is not impacted by distortion and therefore is more accurate than spherical warping that only considers RGB image~\cite{Gardner2017}.
}
\label{fig:warping}
\end{figure} 

\subsection{Point Cloud Generation}
\label{subsec:point_cloud_gen}

The transformation module takes an RGB-D image, represented as $(C, D)$, and the rendering location $r$ and outputs a point cloud of the rendering location. 
Our key insights for using point cloud data format are two-folds:
\1 point cloud can effectively represent the environment geometry and support view transformation;
\2 point cloud resembles the \mci optimization used for real-time \sh calculation.

Figure~\ref{fig:warping} compares the warped results between traditional sphere warping~\cite{Gardner2017} and our point cloud based approach. 
Our approach achieved better warping effect by circumventing the distortion problem associated with the use of RGB images. 
More importantly, our point cloud transformation only requires a simple linear matrix operation, and therefore has the promise to achieve fast lighting estimation for heterogeneous mobile hardware, during inference. 

\begin{figure}[t]
\centering
\includegraphics[width=0.95\textwidth]{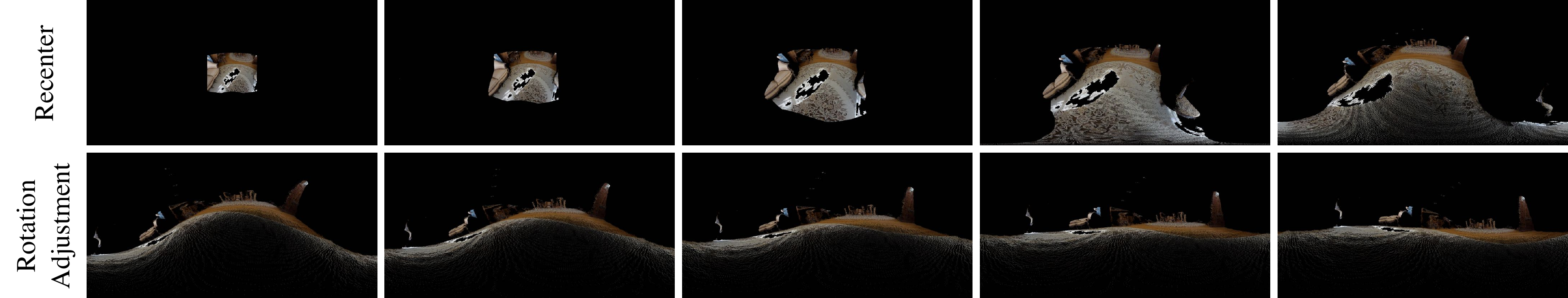}
\caption{\textbf{Example video frames showing point cloud transformation process.}
Row 1 shows the recenter process that simulates a linear camera movement in 3D space. 
Row 2 shows the rotation adjustment on the recentered point cloud.}
\label{fig:recenter}
\end{figure}

Figure~\ref{fig:recenter} shows example video frames when transforming an RGB-D image at observation position by recentering and rotating the generated point cloud, to the rendering position. 
We detail the process of generating point cloud and its corresponding \SHc from a large-scale real-world indoor dataset in Section~\ref{subsec:dataset_generation}.

\subsection{From Point Cloud to SH Coefficients Estimation}
\label{subsec:shc_estimation}

Our second component takes the point cloud $P_r$ at rendering position $r$ and estimates the \SHc $S_r$ which is a compact representation of lighting information at location $r$. 
Our choice of learning \SHc $S_r$ \emph{directly}, instead of other representations such as image style irradiance map, is largely driven by our design goal, i.e,. efficient rendering in commodity mobile phones. 
Formulating the illumination as a pixel-wise regression problem~\cite{Song2019} often requires complex neural network designs. As mobile augmented reality applications often have a tight time budget, e.g., 33 ms for 30fps UI update, it can be challenging to support the use of such neural networks directly on mobile devices. Additionally, popular rendering engines such as Unreal Engine support rendering 3D objects directly with \SHc. 

To train $h: h(P_r, r) = S_r$, we chose the
PointConv~\cite{Wu_2019_CVPR} model, an efficient implementation for building deep convolutional networks directly on 3D point clouds. It uses multi-layer perceptrons to approximate convolutional filters by training on local point coordinates.

This component is trained with supervision from a \SHc $l2$ loss $L_{S}$ as defined in Equation~\eqref{eq:loss}, similar to \ctgaron. 

\begin{align} 
L_{S} = \frac{1}{9}\sum^{3}_{c=1}\sum^{2}_{l=0}\sum^{l}_{m=-l}(i^{m*}_{l,c} - i^{m}_{l,c}),\label{eq:loss} 
\end{align}

\noindent where $c$ is the color channel (RGB), $l$ and $m$ are the degree and order of \SHc. 
We chose to target 2nd order \SHc as it is sufficient for our focused application scenarios, i.e., diffuse irradiance learning~\cite{ramamoorthi_efficient_2001}. We envision that \sysname will run in tandem with existing techniques such as environment probe in modern mobile AR frameworks to support both specular and diffuse material rendering.

\begin{figure}[t]
\centering
\includegraphics[width=0.95\textwidth]{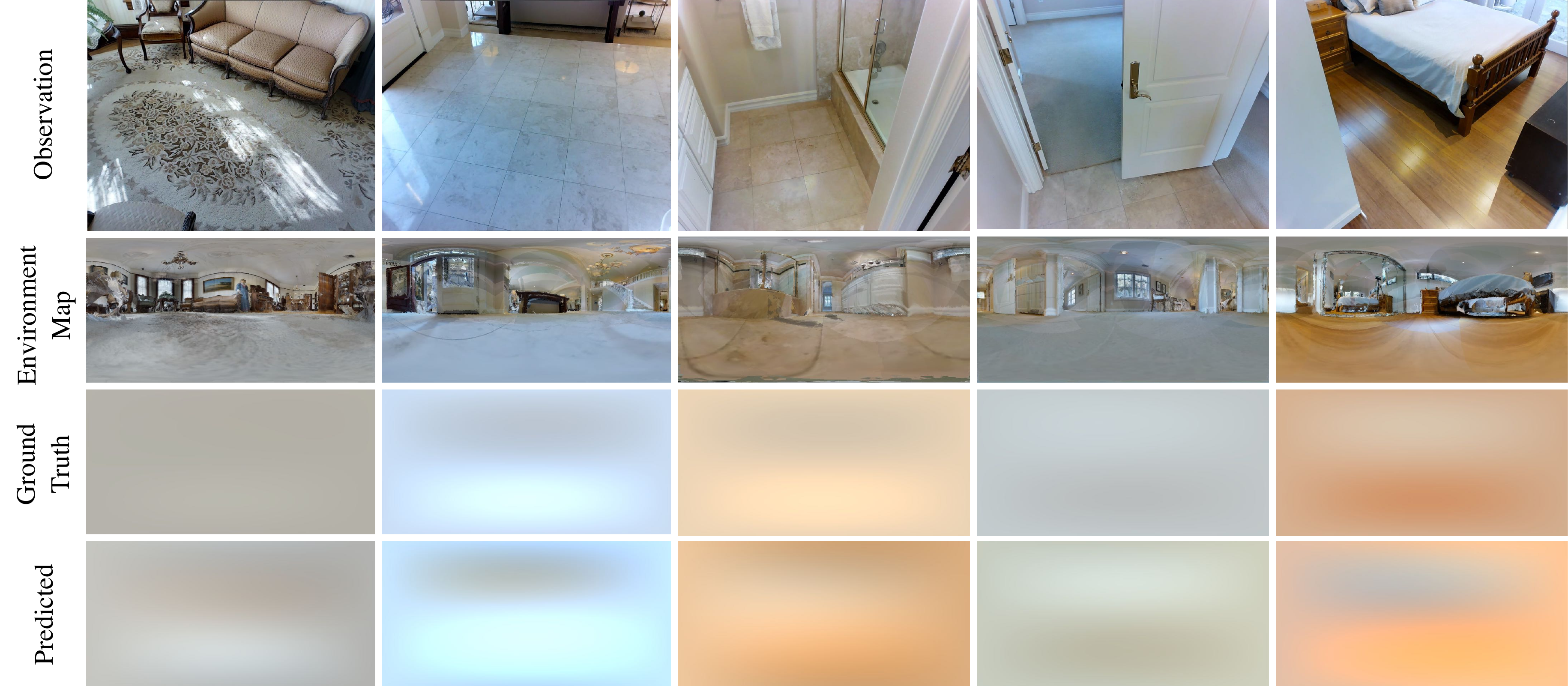}
\caption{\textbf{Irradiance map comparison between ground truth and predicted ones.} 
Row 1 shows the observation captured by mobile camera. 
Row 2 shows environment map generated from the dataset. 
Row 3 shows the irradiance map generated from the environment map using spherical harmonics convolution. 
Row 4 shows the reconstructed irradiance map from \SHc predicted by \sysname.}
\label{fig:irradiance}
\end{figure}

\subsection{Dataset Generation of Point Clouds and SH Coefficients}
\label{subsec:dataset_generation}

Next, we describe how we generated a large-scale real-world training dataset by leveraging two existing datasets, i.e., Matterport3D~\cite{chang2017matterport3d} and Neural Illumination~\cite{Song2019} datasets.
Briefly, Matterport3D contains 194,400 RGB-D images forming 10,800 panoramic views for indoor scenes. 
Each RGB-D panorama contains aligned color and \emph{depth} images (of size 1028x1024) for 18 viewpoints. 
The Neural Illumination dataset was derived from Matterport3D and contains additional relationship between images at observation and rendering locations.

The first step of our dataset creation process is to transform the observation RGB-D images, i.e., captured at user locations, in Matterport3D dataset into point cloud format. 
To do so, we leveraged the pinhole camera model~\cite{chuang2005camera} and camera intrinsics of each photo in the dataset.
For each RGB-D photo, we first recovered a small portion of missing depth data by preprocessing the corresponding depth image with the cross bilateral filter. Then we calculated the 3D point cloud coordinates $(x, y, z)$ as: 
\begin{align} 
x = \frac{(u - cx) * z}{fx}, \quad y = \frac{(v - cy) * z}{fy}, \label{eq:pointCloudTransformation} \nonumber
\end{align}

\noindent where $z$ is the depth value in the RGB-D photo, $u$ and $v$ are the photo pixel coordinates, $fx$ and $fy$ are the vertical and horizontal camera focal length, $cx$ and $cy$ are the photo optical center. 

With the above transformation, we generated the point cloud $P_{o}$ for each observation position. Then, we applied a linear translation $T$ to $P_{o}$ to transform the view at observation position to the rendering position. 
Specifically, $T$ is determined by using the pixel coordinates of each rendering position on observation image from the Neural Illumination dataset in order to calculate a vector to the locale point. To represent the rendering position for a 3D object, we used a scale factor. This allows us \1 to compensate for the position difference between the placement and the ground truth locations; \2 to account for the potentially inaccurate depth information that was introduced by IR sensors. Currently, we used 0.95 based on empirical observation of point cloud projection.

We also used a rotation operation that aligns the recentered point cloud $P_{o}$ with ground truth environment maps in our dataset. Both the recenter and rotation operations are needed during the inference to achieve spatially-variant lighting estimation and to account for the geometry surface and camera pose.

Finally, for each panorama at the rendering position, we extracted 2nd order \SHc as 27 float numbers to represent the irradiance ground truth. 
On generated point clouds, we also performed projection and consequently generated respective 2D panorama images which will be used in the ablation study.

\subsection{Design and Training Discussions}
\label{subsec:discussion}

\subsubsection{Learning From Point Cloud} 
Our choices to learn lighting information \emph{directly} from point cloud and estimating diffuse environment map are driven by mobile-specific challenges and inference time requirement. For example, it is challenging to construct detailed environment map from a limited scene view captured by the mobile phone camera. This in turns can lead to more complex neural networks that might not be suitable to run on mobile devices. Furthermore, neural network generated environment maps may be subject to distortion and unexpected shape. This might lead to reflective textures during rendering and can significantly affect the AR end-user experience.

One intuitive approach is to formulate the learning process as an image learning task by projecting the transformed point cloud into a panorama. 
This is because image-based lighting models~\cite{debevec2006image} commonly use $360^{\circ}$ panoramic view to calculate lighting from every direction of the rendering position.  
However, learning from projected point cloud can be challenging due to potential image distortion and missing pixel values. 
We compare this formulation to \sysname in Section~\ref{sec:eval}.

Our idea to learn diffuse lighting from the point cloud representation is in part inspired by how real-time diffuse lighting calculation has been optimized in modern 3D rendering engines. 
Ramamoorthi et al. proposed the use of \sh convolution to speedup the irradiance calculation~\cite{ramamoorthi_efficient_2001}. Compared to diffuse convolution, \sh convolution is approximately $O(\frac{T}{9})$ times faster where $T$ is the number of texels in the irradiance environment map \cite{ramamoorthi_efficient_2001}.
However, as \sh convolution still includes integral operation, performing it directly on large environment maps might hurt real-time performance. 
Consequently, \mci was proposed as an optimization to speed up lighting calculation through \sh convolution by uniformly sampling pixels from the environment map. 

In short, \mci demonstrates the feasibility to calculate the incoming irradiance with enough uniformly sampled points of the environment map. 
In our problem setting of mobile AR, we have limited samples of the environment map which makes it nature to formulate as a data-driven learning problem with a neural network. 

Directly learning from point cloud representation can be difficult due to the sparsity of point cloud data~\cite{liu2019deep}. We chose a recently proposed PointConv~\cite{Wu_2019_CVPR} architecture as an example point cloud-based neural network for lighting estimation (the second stage of \sysname). Other point cloud learning approaches might also be used~\cite{li2018pointcnn,qi2016pointnet,xu2018spidercnn}. 

\subsubsection{Training Dataset} 
Training a neural network to accurately estimate lighting requires a large amount of real-world indoor 3D scenes that represent complicated indoor geometries and lighting variances.
Furthermore, in the context of mobile AR, each training data item needs to be organized as a tuple of 
\1 a RGB-D photo $(C, D)$ captured by mobile camera at the observation position to represent the user's observation; 
\2 a $360^{\circ}$ panorama $E$ at the rendering position for extracting lighting information ground truth;
\3 a relation $R$ between $(C, D)$ and $E$ to map the pixel coordinates at the observation location to the ones at the rendering position, as well as the distance between these pixel coordinates. 

Existing datasets all fall short to satisfy our learning requirements~\cite{chang2017matterport3d,2017arXiv170201105A,Song2019}. For example, Matterport3D provides a large amount of real-world panorama images which can be used to extract lighting information to serve as ground truth. However, this dataset does not include either observation nor relation data. The Neural Illumination dataset has the format that is closest to our requirements but is still missing information such as point clouds. 

In this work, we leveraged both Matterport3D and the Neural Illumination datasets to generate a dataset that consists of point cloud data $P$ and \SHc $S$. Each data entry is a five-item tuple represented as $((C,D), E, R, P, S)$. 
However, training directly on the point clouds generated from observation images are very large, e.g., 1310720 points per observation image, which complicates model training with can be inefficient as each observation image contains 1310720 points, requiring large amount of GPU memory. 
In our current implementation, we uniformly down-sampled each point cloud to 1280 points to reduce resource consumption during training and inference. 
Our uniform down-sampling method is consistent with the one used in the PointConv paper~\cite{Wu_2019_CVPR}.
Similar to what was demonstrated by Wu et al.~\cite{Wu_2019_CVPR}, we also observed that reducing the point cloud size, i.e., the number of points, does not necessarily lead to worse prediction but can reduce GPU memory consumption linearly during training. 
\section{Evaluation}
\label{sec:eval}

\begin{figure}[t]
\centering
\includegraphics[width=0.95\textwidth]{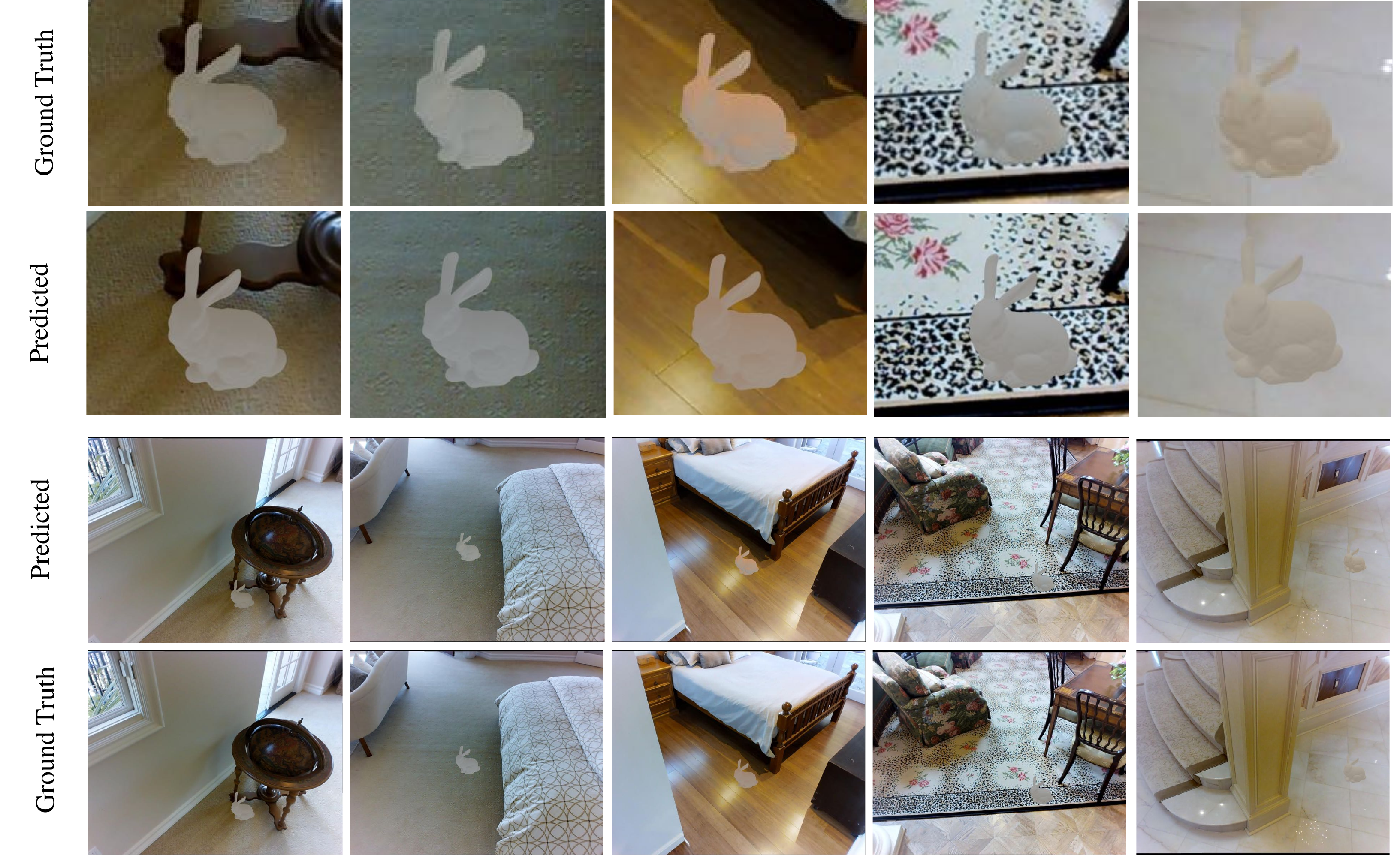}
\caption{\textbf{Rendering example from \sysname.} Comparison between rendering a 3D object with ground truth irradiance map and irradiance map generated by \sysname. 
Row 1 and 2 show the comparison between a closeup look of the rendered objects. 
Row 3 and 4 show the comparison between rendered objects in original observations. 
Note we did not cast shadow to highlight lighting-related rendering effects to avoid misleading visual effects.}
\label{fig:rendering}
\end{figure}

We trained and tested \sysname on our generated dataset (Section~\ref{subsec:dataset_generation}) by following the train/test split method described in previous work~\cite{Song2019}. 
Our evaluations include end-to-end comparisons to recent works on lighting estimation, ablation studies of our pipeline design, and resource complexities compared to commonly used mobile DNNs.

\subsubsection{Evaluation Metrics} 

We use the following two metrics to quantitatively evaluate the accuracy on our predicted \SHc $S_r$: \1 \emph{\SHc $l2$ distance loss} is the average of all \SHc $l2$ distance, which is calculated as the numerical difference between the $S_r$ predicted by \sysname and the ground truth. \2 \emph{Reconstructed irradiance map $l2$ distance loss} is defined as the average of pixel-wise $l2$ distance loss on reconstructed irradiance map. In our \sysname pipeline, we need to first reconstruct an irradiance map (see Equation 7 in Ramamoorthi et al. \cite{ramamoorthi_efficient_2001}) and then compare to the ground truth extracted directly from our dataset, to calculate the $l2$ loss. We compare this metric to the diffuse loss, defined by \ctsong, as both representing the reconstruction quality of irradiance map.

\subsubsection{Hyperparameter Configurations} We adopted a similar model architecture used by \ctpointconv for ModelNet40 classification~\cite{wu20153d}.
We used a set of hyperparameters, i.e., 2 PointConv blocks, each with multilayer perceptron setup of (64, 128) and (128, 256), based on accuracy and memory requirement. 

\setlength{\tabcolsep}{4pt}
\begin{table}[t]
\caption{\textbf{Comparison to state-of-the-art networks.} Our approach \sysname (highlighted in green) achieved the lowest loss for both \shc $l2$ and irradiance map $l2$.  Note Song et al. used traditional diffuse convolution to generate irradiance map and did not use \SHc.}
\label{table:compare_state_of_the_art}
\centering
\scriptsize{
\begin{tabular}{lrr}
\toprule
\textbf{Method} & \textbf{\SHc $l2$ Loss} & \textbf{Irradiance Map $l2$ Loss}\\
\midrule
\ctsong & N/A & 0.619 \\
\ctgaron & 1.10 ($\pm$ 0.1) & 0.63 ($\pm$ 0.03) \\
\textbf{\sysname (Ours)} & \textbf{0.633 ($\pm$ 0.03)} & \textbf{0.433 ($\pm$ 0.02)}\\
\bottomrule
\end{tabular}
}
\end{table}
\setlength{\tabcolsep}{1.4pt}

\subsubsection{Comparisons to State-of-the-art} We preformed quantitative comparison experiments with two state-of-the-art end-to-end deep learning model architectures: \1 Song et al.~\cite{Song2019}; and \2 Garon et al.~\cite{Garon2019}. Table~\ref{table:compare_state_of_the_art} shows the comparison on two loss metrics.

Song et al. estimates the irradiance by using a neural network pipeline that decomposes the lighting estimation task into four sub-tasks: 
\1 estimate geometry, \2 observation warp, \3 LDR completion, and \4 HDR illumination. 
As we used the same dataset as Song et al., we obtained the corresponding irradiance map $l2$ loss from the paper. However, since Song et al. used the traditional diffuse convolution to obtain irradiance map, the paper did not include \SHc $l2$ loss. 
Garon et al. estimates the \SHc represented lighting and a locale pixel coordinate of a given input image by training a two-branch convolutional neural network with end-to-end supervision.
We reproduced the network architecture and trained on our dataset, excluding point clouds and relation $E$.  

Table~\ref{table:compare_state_of_the_art} shows that our \sysname achieved 31.3\% and 30\% lower irradiance map $l2$ loss compared to Garon et al. and Song et al., respectively.
We attribute such improvement to \sysname's ability in handling spatially variant lighting with effective point cloud transformation. 
Further, the slight improvement (1.7\%) on irradiance map $l2$ loss achieved by Song et al. over Garon et al. is likely due to the use of depth and geometry information.

\setlength{\tabcolsep}{4pt}
\begin{table}[t]
\centering
\caption{\textbf{Comparison to variants.} Row 1 and 2 compare the lighting estimation accuracy with two input formats: point cloud and projected point cloud panorama image. 
Row 3 to 6 compare the lighting estimation accuracy with different downsampled input point clouds.}
\label{table:compare_variants}
\scriptsize{
\begin{tabular}{lrr}
\toprule
\textbf{Method} & \textbf{\SHc $l2$ Loss} & \textbf{Irradiance Map $l2$ Loss}\\
\midrule
Projected Point Cloud + ResNet50  &0.781 ($\pm$ 0.015) & 0.535 ($\pm$ 0.02)\\
\textbf{\sysname (Point Cloud + PointConv)} & \textbf{0.633 ($\pm$ 0.03)} & \textbf{0.433 ($\pm$ 0.02)} \vspace{1.5mm} \\
512 points & 0.668 ($\pm$ 0.02) & 0.479 ($\pm$ 0.02)\\
768 points  & 0.660 ($\pm$ 0.02) & 0.465 ($\pm$ 0.02)\\
1024 points & 0.658 ($\pm$ 0.03) & 0.441 ($\pm$ 0.02)\\
\textbf{1280 points (\sysname)} & \textbf{0.633 ($\pm$ 0.03)} & \textbf{0.433 ($\pm$ 0.02)}\\
\bottomrule
\end{tabular}
}
\end{table}
\setlength{\tabcolsep}{1.4pt} 

\subsubsection{Comparisons to Variants} 
To understand the impact of neural network architecture on the lighting estimation, we further conducted two experiments: \1 learning from projected point cloud and \2 learning from point clouds with different number of points. Table~\ref{table:compare_variants} and Table~\ref{table:compare_complexity} compare the model accuracy and complexity, respectively. 

In the first experiment, we study the learning accuracy with two different data representations, i.e., projected point cloud and point cloud, of 3D environment. We compare learning accuracy between learning from point cloud directly with PointConv and learning projected point cloud panorama with ResNet50, which was used in \ctsong. 
In this experiment, we observed that learning from projected point cloud resulted in lower accuracy (i.e., higher $l2$ losses) despite that ResNet50 requires an order of magnitude more parameters (i.e., memory) and MACs (i.e., computational requirements) than PointConv. The accuracy difference is most likely due to the need to handle image distortion, caused by point cloud projection. 
Even though there might be other more suitable convolution kernels than the one used in ResNet50 for handing image distortion, the high computational complexity still makes them infeasible to directly run on mobile phones. 
In summary, our experiment shows that learning lighting estimation from point cloud achieved better performance than traditional image-based learning.

In the second experiment, we evaluate the performance difference on a serial of down-sampled point clouds. 
From Table~\ref{table:compare_complexity}, we observe that the multiply accumulates (MACs) decreases proportionally to the number of sampled points, while parameters remain the same. 
This is because the total parameters of convolution layers in PointConv block do not change based on input data size, while the number of MACs depends on input size. 
Furthermore, we observe comparable prediction accuracy using down-sampled point cloud sizes to \sysname, as shown in Table~\ref{table:compare_variants}. 
An additional benefit of downsampled point cloud is the training speed, as less GPU memory is needed for the model and larger batch sizes can be used.
In summary, our results suggest the potential benefit for carefully choosing the number of sampled points to trade-off goals such as training time, inference time, and inference accuracy. 

\setlength{\tabcolsep}{4pt}
\begin{table}[t]
\caption{\textbf{Comparison of model complexities.} Row 1 to 4 compare resource complexity of ResNet50 (which is used as one component in \ctsong) and mobile-oriented DNNs to that of \sysname.  
Row 5 to 8 compare the complexity with different downsampled input point clouds.}
\label{table:compare_complexity}
\centering
\scriptsize{
\begin{tabular}{lrr}
\toprule
\textbf{Model} & \textbf{Parameters(M)} & \textbf{MACs(M)} \\
\midrule
ResNet50~\cite{He2015} & 25.56 & 4120 \\
MobileNet v1 1.0\_224~\cite{howard2017mobilenets} & 4.24 & 569 \\
SqueezeNet 1\_0~\cite{iandola2016squeezenet} & 1.25 & 830 \\
\textbf{\sysname (Ours)} & \textbf{1.42} & \textbf{790}\vspace{1.5mm}\\
512 points & 1.42 & 320 \\
768 points & 1.42 & 470 \\
1024 points & 1.42 & 630\\
\textbf{1280 points (\sysname)} & \textbf{1.42} & \textbf{790}\\
\bottomrule
\end{tabular}
}
\end{table}
\setlength{\tabcolsep}{1.4pt}
\vspace{-6mm}

\subsubsection{Complexity Comparisons} To demonstrate the efficiency of our point cloud-based lighting estimation approach, we compare the resource requirements of \sysname to state-of-the-art mobile neural network architectures~\cite{howard2017mobilenets,iandola2016squeezenet}.  
We chose the number of parameters and the computational complexity as proxies to the inference time~\cite{Sze2017survey}.
Compared to the popular mobile-oriented models MobileNet~\cite{howard2017mobilenets}, our \sysname only needs about 33.5\% memory and 1.39X of multiple accumulates (MACs) operations, as shown in Table~\ref{table:compare_complexity}. 
Further, as MobileNet was shown to produce inference results in less than 10ms~\cite{tensorflow_mobile_models}, it indicates \sysname's potential to deliver real-time performance. Similar observations can be made when comparing to another mobile-oriented model SqueezeNet~\cite{iandola2016squeezenet}.   
\section{Related Work}
\label{sec:related}

Lighting estimation has been a long-standing challenge in both computer vision and computer graphics.
A large body of work~\cite{Gardner_2019_ICCV,Gardner2017,gruber2012real,Song2019,zhang2018discovering} has been proposed to address various challenges and more recently for enabling real-time AR on commodity mobile devices~\cite{Garon2019}.

\subsubsection{Learning-based Approaches}
Recent works all formulated the indoor lighting estimation problem by learning directly from a single image, using end-to-end neural networks~\cite{Gardner2017,Garon2019,Song2019}. 
Gardner et al. trained on an image dataset that does not contain depth information and their model only outputs one estimate for one image~\cite{Gardner2017}. Consequently, their model lacks the ability to handle spatially varying lighting information~\cite{Gardner2017}.  
Similarly, Cheng et al. proposed to learn a single lighting estimate in the form of \SHc for an image by leveraging both the front and rear cameras of a mobile device~\cite{ChengSCDZ18_graph}.
In contrast, Garon et al. proposed a network with a global and a local branches for estimating spatially-varying lighting information by training on indoor RGB images~\cite{Garon2019}. However, the model still took about 20ms to run on a mobile GPU card and might not work for older mobile devices. 
\ctsong proposed a fully differential network that consists of four components for learning respective subtasks, e.g., 3D geometric structure of the scene. Although it was shown to work well for spatially-varying lighting, this network is too complex to run on most of the mobile devices.
In contrast, our \sysname not only can estimate lighting for a given locale (spatially-variance), but can do so quickly with a compact point cloud-based network. 
Further, our work generated a dataset of which each scene contains a dense set of observations in the form of (point cloud, \SHc).

\subsubsection{Mobile AR}
Companies are providing basic support for developing AR applications for commodity mobile devices in the form of development toolkits~\cite{arcore_website,arkit_website}.
However, to achieve seamless AR experience, there are still a number of mobile-specific challenges. For example, it is important to detect and track the positions of physical objects so as to better overlay the virtual 3D objects~\cite{Apicharttrisorn2019-fa,Liu_undated-pn,Tulloch2018-wj}. 
Apicharttrisorn et al.~\cite{Apicharttrisorn2019-fa} proposed a framework that achieves energy-efficiency object detection by only using DNNs as needed, and leverages lightweight tracking method opportunistically. Due to its impact on visual coherence, lighting estimation for AR has received increasing attention in recent years~\cite{Apple_undated-es,google_arcore,Prakash2019-gb}. GLEAM proposed a non-deep learning based mobile illumination estimation framework that relies on physical light probes~\cite{Prakash2019-gb}.
Our work shares similar performance goals, i.e., being able to run AR tasks on mobile devices, and design philosophy, i.e., by reducing the reliance on complex DNNs. 
Unlike prior studies, our work also focuses on rethinking and redesigning lighting estimation, an important AR task for realistic rendering, by being mobile-aware from the outset.

\section{Conclusion and Future Work}

In this work, we described a two-stage lighting estimation pipeline \sysname that consists of an efficient mathematical model and a compact deep learning model. \sysname provides spatially-variant lighting estimation in the form of \SHc at any given 2D locations of an indoor scene.

Our current focus is to improve the lighting estimation accuracy for each camera view captured by mobile devices, given the real-time budgets.
However, mobile AR applications need to run on heterogeneous resources, e.g., lack of mobile GPU support, and have different use cases, e.g., 60fps instead of 30fps, which might require further performance optimizations. As part of the future work, we will explore the temporal and spatial correlation of image captures, as well as built-in mobile sensors for energy-aware performance optimization.

\subsubsection{Acknowledgement}
This work was supported in part by NSF Grants \#1755659 and \#1815619.

\bibliographystyle{src/splncs04}
\bibliography{egbib}

\newpage

\end{document}